\documentclass{svproc}
\usepackage{url}
\usepackage[utf8]{inputenc} 
\usepackage[T1]{fontenc}    
\usepackage{hyperref}       
\usepackage{booktabs}       
\usepackage{amsfonts}       
\usepackage{nicefrac}       
\usepackage{microtype}      
\usepackage{times}
\usepackage{latexsym}
\usepackage{graphicx}
\usepackage{enumitem}
\usepackage{array}
\usepackage[dvipsnames]{xcolor}
\usepackage{graphics}
\usepackage{wrapfig}
\usepackage{mathtools}

\begin{document}
\mainmatter              
\title{Topic Segmentation of Semi-Structured and Unstructured Conversational Datasets using Language Models}
\titlerunning{Topic Segmentation for Conversational data}  
%
\author{Reshmi Ghosh\inst{1} \and Harjeet Singh Kajal\inst{2}
Sharanya Kamath\inst{2} \and Dhuri Shrivastava\inst{2} \and Samyadeep Basu\inst{3} \and Hansi Zeng\inst{2} \and Soundararajan Srinivasan\inst{1}}

\authorrunning{Reshmi Ghosh et al.} 

%
\institute{Microsoft Corp., Cambridge, MA 02142, USA,\\
\and
University Of Massachusetts,
Amherst, MA 01002, USA,
\and
University of Maryland, 
College Park, MD 20742, USA}

\maketitle              

\begin{abstract}
Breaking down a document or a conversation into multiple contiguous segments based on its semantic structure is an important and challenging problem in NLP, which can assist many downstream tasks. However, current works on topic segmentation often focus on segmentation of structured texts. In this paper, we comprehensively analyze the generalization capabilities of state-of-the-art topic segmentation models on unstructured texts. We find that: (a) Current strategies of pre-training on a large corpus of structured text such as Wiki-727K \textit {do not help} in transferability to unstructured conversational data. (b) Training from scratch with only a relatively small-sized dataset of the target unstructured domain improves the segmentation results by a significant margin. We stress-test our proposed Topic Segmentation approach by experimenting with multiple loss functions, in order to mitigate effects of imbalance in unstructured conversational datasets. Our empirical evaluation indicates that Focal Loss function is a robust alternative to Cross-Entropy and re-weighted Cross-Entropy loss function when segmenting unstructured and semi-structured chats.
\end{abstract}
\section{Introduction}
Topic Segmentation refers to the task of splitting texts into meaningful segments that correspond to a distinct topic or a subtopic. Natural language texts, especially in unstructured formats such as chat conversations and transcripts, often do not have an easy-to-detect separation between contiguous topics. Reliable \& accurate division of text into coherent segments can help in making the text more readable as well as searchable. Hence, topic segmentation enables numerous applications such as search assistance and recommendation \cite{10.5555/647344.724145}. It has also been noted that text segmentation can improve and speedup applications such as information extraction and summarization \cite{koshorek2018text}.

 Historically, Topic Segmentation methods have primarily been dependent on lexical chains and machine learning methods that can detect changes in document structure \cite{Documentstructure}.  Recently, a handful of approaches leveraging language models have been proposed for topic segmentation \cite{koshorek2018text},\cite{glavavs2021training},\cite{lukasik2020text}. However, the datasets on which these approaches have been evaluated are often structured in nature such as Wiki-727K \cite{koshorek2018text},\cite{lee2023topic}, Wiki-50  \cite{koshorek2018text}, RST, and Choi \cite{choi2000advances}. Adding to the constraints, in many applications, texts that need to be segmented are often unstructured such as chat transcripts and conversations. But, understanding the effectiveness of topic segmentation methods on such unstructured texts hasn't been well studied. In this paper, we empirically investigate the effectiveness of various topic segmentation methods on unstructured segmentation datasets such as LDC BOLT chat \cite{BOLT} and Amazon Topical chat \cite{gopalakrishnan2019topical}. In addition to being less structured than the Wiki-727K or Wiki-50 data, these datasets are challenging for traditional topic segmentation approaches as they contain grammatically ill-formed "noisy sentences" and a varying number of segments per conversation.  Hence, we systematically examine the effectiveness of large-scale pre-training, dataset used in pre-training, and fine-tuning strategies on these "out-of-domain" (data that is conversational in nature rather than the segmented Wiki content), unstructured text-segmentation datasets spanning traditional the LSTM-based models and the newer transformer-based architectures. 

We find that large-scale pre-training (and fine-tuning with data from target domain) has only a {\it negligible} effect on downstream segmentation tasks, when the task consists of unstructured data. This is contrary to the conventional wisdom in NLP, where pre-training and fine-tuning is a common practice. We, therefore, identify topic segmentation on unstructured data as one task where large-scale pre-training doesn't have any significant effect. To perform well on segmentation of unstructured text, we find that training, from scratch, with only a {\it few} examples of the segmentation domain is sufficient. This is true for segmentation architectures ranging from the LSTM-based ones to the recent Transformer-based ones.  Our results also show that, for unstructured topic segmentation, avoiding pre-training on a large corpus such as the Wiki-727K dataset results in saving a significant amount of training time and compute resources, facilitating the exploration of newer approaches. In summary, our contributions are as follows:

\begin{itemize}[noitemsep,topsep=0pt,parsep=0pt,partopsep=0pt]
    \item We investigate and stress-test the effectiveness of current topic segmentation methods on unstructured texts, which is a more challenging segmentation task when compared to segmentation tasks based on structured datasets. 
    \item We find that pre-training on large topic segmentation datasets such as Wiki-727K has negligible effect on downstream transfer to unstructured text-segmentation datasets and instantiating the model with only a few-examples of the unstructured task is sufficient. 
    \item We present simple and practical recipes to improve topic segmentation performance on unstructured datasets to remedy the effect of imbalanced class labels synthetically generated for the setup of supervised learning approach. We empirically examine the impact of alternative loss functions like re-weighted cross-entropy loss and focal loss on imbalanced dataset, and conclude that focal loss is a more effective alternative for topic segmentation tasks on conversational datasets.
\end{itemize}

\section{Related Works}\label{related_works}

Topic segmentation has been explored through many realms; particularly, the approaches used could be broadly categorized as Non-neural-based and Neural-based approaches in both supervised and unsupervised \cite{galley2003discourse},\cite{glavas-etal-2016-unsupervised} settings.\\

\textbf{Non-neural approaches}:
The early research efforts related to non-neural\cite{Beeferman99statisticalmodels} approaches for topic segmentation include \cite{TextTiling} that focused on an unsupervised approach to analyze lexical cohesion in small segments by leveraging counts of word repetitions. This work was expanded to enable models to understand words and sentences occurring in segments in a comprehensive manner leading to the wide use of lexical chains \cite{adarve2007topic},\cite{sitbon2007topic},\cite{joty2013topic},\cite{riedl2012topictiling}.\\

\textbf{Neural Approaches:}\label{non_neural} \cite{koshorek2018text} used a hierarchical Bi-LSTM to cast the topic segmentation as a supervised learning task. Other neural methods also leverage the Transformer architecture. In \cite{lukasik2020text}, the authors proposed three transformer-based models, of which Cross-Segment BERT model is particularly important for topic segmentation tasks as the model captures information from the local context surrounding a potential topic segment boundary to judge about which pool of sub-document units, the particular segment belongs. The other two model architectures use a hierarchical approach as used by \cite{koshorek2018text}, but with using the BERT model instead of BiLSTMs. \cite{DBLP:journals/corr/abs-2106-12978} is another recent work that uses an unsupervised approach based on BERT embeddings to segment topics in multi-person meeting transcripts.\\

\section{Segmentation Datasets}\label{data}

\textbf{Unstructured Datasets}: We utilize the Linguistic Data Consortium (LDC) and BOLT SMS/chat data collection (restricted to English chats and SMS, henceforth referred to as BOLT) and Amazon Topical Chat dataset. The BOLT dataset is considerably more `unstructured' compared to the Topical chat dataset, as it comprises non-uniform sentence structures, incomplete sentences, and abbreviations, commonly found in asynchronous conversations on direct messaging applications. Consequently, we refer to BOLT as unstructured dataset and categorize the Topical chat dataset as semi-structured dataset due to its more coherent sentence structure.



\begin{figure}[!ht]
    \includegraphics[width=\textwidth]{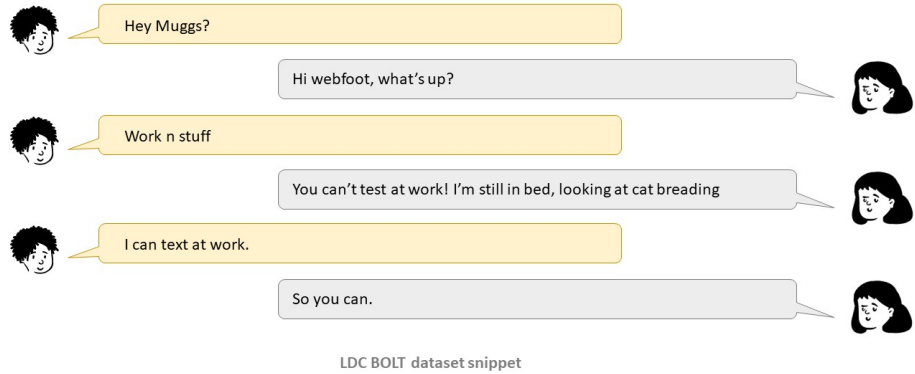}\hfill

    \caption{Snapshot of the LDC Bolt dataset which contains conversations that have a higher degree of 'unstructured-ness', meaning, incomplete sentences, usage of abbreviations, and little or no punctuation. The LDC BOLT dataset more closely represents the modern-day, fast paced SMS/text conversations. }\label{fig:LDC}
\end{figure}

The LDC BOLT SMS and Chat dataset (Figure \ref{fig:LDC}) includes conversations that have been extracted from messaging platforms like WhatsApp, iMessage, Android SMS, Symbian SMS, Viber, BlackBerry, QQ, Google chat, Skype chat, \& Yahoo Messenger in Chinese, Egyptian Arabic, and English. The dataset contains 2140 Egyptian, 7844 Chinese and 9155 English conversations. Our analysis is limited to the English subset of the data.The dataset exhibits the following characteristics:
         \begin{itemize}
         \item Mean sentence length (number of words) = 9.45 ; standard deviation = 9.41
         \item Mean segment length (number of sentences) = 11.28; standard deviation  = 13.99
         \end{itemize}

\begin{figure}[!ht]
    \includegraphics[width=\textwidth]{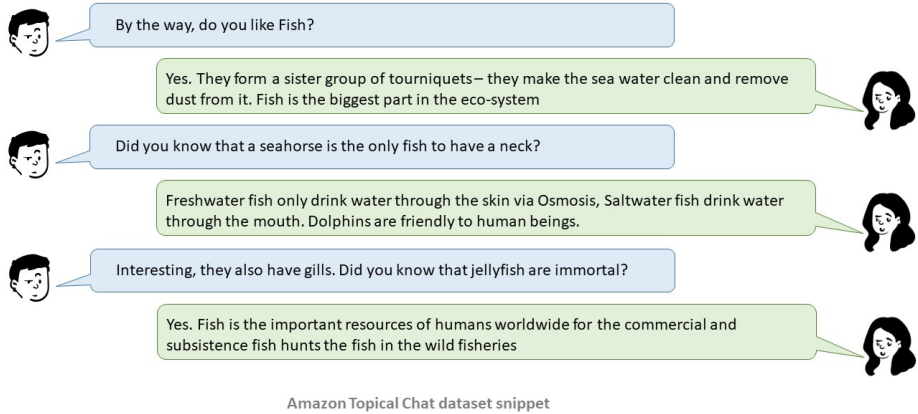}\hfill

    \caption{Snapshot of Amazon Topical Chat dataset, which is a repository of conversations that have well-formed sentences and appropriate punctuation. }\label{fig:Topical}
\end{figure}

The Amazon Topical Chat dataset (Figure \ref{fig:Topical}) contains human-to-human conversations spanning eight broad topics, with over 8000 conversations. Although the dataset features well-structured sentences, it consists of brief conversations rather than articles, prompting its classification as semi-structured. The Topical Chat dataset's underlying knowledge encompasses eight broad topics, with minimal variation in the segment length. 
         \begin{itemize}
         \item Mean sentence length (number of words) = 19.86, standard deviation  = 10.49
         \item Mean segment length (number of sentences) = 21.83, standard deviation  = 1.75
        \end{itemize}

\textbf{Wiki-727K}: Proposed by \cite{koshorek2018text}, Wiki-727K comprises 727,746 English-language documents with text segmentations based on their table of contents. As the text is non-conversational and features proper syntactical structure in the form of well-organized sentences, paragraphs, and sections, the Wiki-727K dataset is deemed structured.


To the best of our knowledge, Wiki-727K is the sole publicly available large dataset suitable for large-scale pre-training \cite{koshorek2018text},\cite{glavavs2021training},\cite{lukasik2020text}. No single conversational (unstructured) dataset of comparable  scale exists for extensive pre-training of large Transformer models. A similar approach of pre-training was employed in \cite{koshorek2018text}\cite{glavavs2021training}\cite{lukasik2020text}. \\


Both Topical chat and BOLT datasets are conversational, necessitating additional pre-processing to adapt them for a supervised learning setup. We accomplish this by segmenting these datasets into multiple segments (using five segments as it yielded the best F1 scores; {\it Figure \ref{fig1}}), with each segment representing a specific chat conversation snippet. After pre-processing the conversations into the appropriate number of segments, we generated synthetic labels for each sentence in every segment, casting the topic segmentation task for conversational datasets as a supervised learning problem with binary labels. We first pre-process the datasets to synthetically create multiple segments, and label the sentences occurring in each segment $x_i = {(x_1, x_2, ...., x_n)}$ based on the boundary condition (i.e., whether a specific sentence was end-of-segment (label, $y_i =$ `1') or a non end-of-segment sentence (encoded label $y_i =$ `0')).

\subsection{Model Architecture}
\label{sec:appendix}
Topic segmentation in the existing literature has involved neural and non-neural approaches. Due to the complexity of understanding heterogeneous conversational datasets, we use state-of-the-art neural models, i.e, Hierarchical Bi-LSTM as proposed by \cite{koshorek2018text}, and CSBERT model as introduced by \cite{lukasik2020text}.\\

The Hierarchical Bi-LSTM model is a neural architecture that first learns sentence representation, which are then fed into a segment prediction sub-network. The lower-level sub-network employs a two-layer bi-directional LSTM layer that generates representations, by consuming words $w\textsubscript{1}, w\textsubscript{2}....w\textsubscript{i}$ of a sentence $x\textsubscript{i}$ as input. The intermediate output is passed through a max pooling layer to create the final sentence representations $e\textsubscript{i}$. The higher-level sub-network for segment prediction takes a sequence of sentence embeddings generated from the lower sub-network, and feeds them into a two-layer Bi-LSTM, which then feeds into a fully connected layer with a softmax function to generate segmentation probabilities.\\

In the Cross-Segment BERT model, the authors leverage information from the local context, i.e., studying the semantic shift in word distributions, as first introduced in \cite{TextTiling}. The additional context on both sides of the segment boundary (termed as `candidate break in the paper), i.e., the sequence of word-piece tokens that come before and after the segment breaks. Basically, the model is fed $k$ word-piece tokens from the left and $k$ tokens from the right of a segment break. The input is composed of a classification token (denoted by $[CLS]$ ), followed by the two contexts concatenated together and separated by a separator token (denoted by $[SEP]$). The tokens are fed into the Transformer encoder(\cite{Vaswani}), which is initialized by $BERT\textsubscript{LARGE}$ to output segmentation probabilities.  The $BERT_\textsubscript{LARGE}$ model has 24 layers and uses 1024-dimensional embeddings and 16-attention heads. As the released BERT checkpoint only supports up to 512 tokens, we keep a maximum 250 word tokens on each side.

The RoBERTa architecture was chosen as a comparative alternative to the BERT model in the Cross-Segment framework, as it is the relatively newer successor of BERT \cite{liu2019roberta}. However, rather than changing the framework of Cross-Segment learning as proposed by \cite{lukasik2020text}, wherein the authors demonstrate the capability of BERT model to learn the context around end-of-segments in a robust way, we chose to utilize the same framework and simply replace the BERT model with RoBERTa-base.

\subsection{Setup}
\label{appdx:exp_setup}

The Wiki-727K dataset was randomly partitioned in 80\% / 10\% / 10\% format to create train, development (fine-tuning), and test set, respectively. We used the partitioned train set to perform pre-training using Hierarchical Bi-LSTM, Cross-Segment BERT, and Cross-Segment RoBERTa models for the first set of experiments, i.e., to evaluate the effectiveness of large-scale pretraining on structured dataset.\\ 

Additionally we synthetically partitioned these conversational datasets to generate segments (snippets of conversations) and group the segment chunks to form documents. The model at any point of the training process consumed a batch of these documents. As described in section \ref{data}, the sentences in these segments are synthetically labelled to  indicate the end-of-segment. As a result, the chunking the conversations from BOLT and Topical Chat datasets in 5 segments leads to 1815 and 1726 documents, respectively. \\

Furthermore, the documents created from the chunking process described above were split into train/fine-tuning/test sets for tasks described in Table \ref{tab:exp4} and Table \ref{tab:exp10}. The documents from Topical Chat dataset was divided in 1099 / 348 / 281 documents for the train/fine-tuning/test splits respectively, wherein each document had 5 segments of mean segment length of approximately 11.5 sentences. Additionally, a similar approach was employed for the documents from BOLT dataset, and it was divided in 1090 / 363 / 362 documents for the train/fine-tuning/test splits respectively containing 5 segments each (mean segment length of ~21.83 sentences). \\


Lastly to train the three models from scratch using the unstructured, and semi-structured data, without involving any pre-training on structured Wiki-727K, required 1109 documents of 5 segments from BOLT, and 1030 documents of 5 segments from Topical Chat. 


\section{Experiments}\label{experiments}

We study the problem of segmenting semi-structured and unstructured chats using three popular modeling paradigms used in the structured segmentation domain: the Hierarchical Bi-LSTM model proposed by \cite{koshorek2018text} and the Cross-Segment BERT model \cite{lukasik2020text} (hereafter CSBERT), and the Cross-Segment RoBERTa (CSRoBERTa)  \cite{glavavs2021training} (Section\ref{sec:appendix}). In CSRoBERTa, we use the same training paradigm as CSBERT, but replace BERT with RoBERTa-base \cite{liu2019roberta}. 

We cast the task of topic segmentation as a binary classification problem, and for the purpose of validating our proposed models, we use Precision, Recall, and F1 scores to measure performance. Precision measures the percentage of boundaries identified by the model that are true boundaries. Complementary to Precision, Recall measures the percentage of true boundaries identified by the model. Although comprehensive, it is important to note that there are some challenges associated with individually reporting Precision and Recall as they are somewhat less sensitive to near misses of true boundary identifications, when the prediction is off by one or two sentences. Hence, we additionally report F1 scores in Section \ref{experiments}. F1 score can be reliably used to conclude our initial findings from performing topic segmentation based binary classification on unstructured and semi-structured data.

Note that Topic Segmentation models in the existing body of work have not been validated against any form of semi-structured or unstructured datasets. Table \ref{tab:exp4} presents the results of our analysis. We first pre-train all three models on Wiki-727K and test against the unstructured BOLT, the semi-structured Topical Chat, and the structured Wiki-727K datasets (underlined in Table \ref{tab:exp4}). We then use different pre-training and fine-tuning combinations to examine the necessity of large-scale pre-training on structured datasets for segmenting conversational data. Also note that the results in Table \ref{tab:exp4} are generated by partitioning the BOLT and Topical Chat data into documents of 5 segments each (validated in Figure \ref{fig1}; more details on train-test split are in Section \ref{appdx:exp_setup}).

Adapting the hierarchical Bi-LSTM model on the unstructured BOLT and the semi-structured Topical chat datasets during the validation phase, we conclude that the F1 scores (additional details on evaluation metrics in section \ref{experiments}) are significantly worse when compared with the performance of the CSBERT and CSRoBERTa model in the same setting. However, evaluating the performance on the structured Wiki-727K dataset, we find that the F1 scores from both models are in the same range. 

\begin{center}
\begin{table*}[h]

\scalebox{0.9}{
\hspace{-1.5cm}
\begin{tabular} {  c|c|c|c|c|c|c}
\hline
&

\multicolumn{3}{c|}{\small \textbf{Datasets}}

&
&
&
\\
\hline
\footnotesize \textbf{Task} 
&
\footnotesize \textbf{Pre-Train} 
&
\footnotesize \textbf{Finetune}
&
\footnotesize \textbf{Test}
&
\footnotesize \textbf{Cross Segment BERT}
&
\footnotesize \textbf{Cross Segment RoBERTa-Base}
&
\footnotesize \textbf{Hierarchical Bi-LSTM}
\\
\hline
\footnotesize A.1 & \footnotesize Wiki-727K &  \footnotesize - & \footnotesize Topical Chat & \footnotesize 0.492 & \footnotesize 0.487 &  \footnotesize 0.021\\
\footnotesize A.2 &\footnotesize Wiki-727K & \footnotesize BOLT & \footnotesize Topical Chat & \footnotesize 0.470 & \footnotesize 0.406 & \footnotesize 0.391 \\
\footnotesize A.3 & \footnotesize Wiki-727K & \footnotesize Topical Chat & \footnotesize Topical Chat & \footnotesize \underline{0.725} & \footnotesize \underline{0.713} & \footnotesize \underline{0.931} \\
\hline
\footnotesize A.4 & \footnotesize BOLT & \footnotesize - & \footnotesize Topical Chat & \footnotesize 0.491 & \footnotesize  0.498 & \footnotesize 0.611\\
\footnotesize A.5 & \footnotesize BOLT & \footnotesize Topical Chat & \footnotesize Topical Chat & \footnotesize 0.734 & \footnotesize 0.729 & \footnotesize 0.915\\
\hline
\footnotesize A.6 & \footnotesize Topical Chat & \footnotesize - &  \footnotesize Topical Chat &  \footnotesize 0.764 &   \footnotesize  \bf\textcolor{purple}{0.767} &  \footnotesize  \bf\textcolor{purple}{0.951}\\
\footnotesize A.7 & \footnotesize Topical Chat & \footnotesize BOLT & \footnotesize Topical Chat & \footnotesize  \bf\textcolor{purple}{0.767} & \footnotesize 0.759 & \footnotesize 0.501\\
\hline
\hline

\footnotesize B.1 & \footnotesize Wiki-727K &  \footnotesize - & \footnotesize BOLT & \footnotesize 0.487 & \footnotesize 0.467 &  \footnotesize 0.005\\
\footnotesize B.2 & \footnotesize Wiki-727K & \footnotesize BOLT & \footnotesize BOLT & \footnotesize \underline{0.489} & \footnotesize \footnotesize \underline{0.479} & \footnotesize \underline{0.406} \\
\footnotesize B.3 & \footnotesize Wiki-727K & \footnotesize Topical Chat & \footnotesize BOLT & \footnotesize 0.511 & \footnotesize 0.492 & \footnotesize 0.152 \\
\hline
\footnotesize B.4 & \footnotesize BOLT & \footnotesize - & \footnotesize BOLT & \footnotesize \bf\textcolor{cyan}{0.569} & \footnotesize \bf\textcolor{cyan}{0.561} & \footnotesize  \bf\textcolor{cyan}{0.443}\\
\footnotesize B.5 & \footnotesize BOLT & \footnotesize Topical Chat & \footnotesize BOLT & \footnotesize 0.518 & \footnotesize 0.509 & \footnotesize 0.181\\
\hline
\footnotesize B.6 & \footnotesize Topical Chat & \footnotesize - &  \footnotesize BOLT &  \footnotesize 0.544 &  \footnotesize  0.542 &  \footnotesize 0.157\\
\footnotesize B.7 & \footnotesize Topical Chat & \footnotesize BOLT & \footnotesize BOLT & \footnotesize 0.536 & \footnotesize 0.529 & \footnotesize 0.331\\

\hline

\hline

\hline
\footnotesize C.1 & \footnotesize Wiki-727K &  \footnotesize - & \footnotesize Wiki-727K & \footnotesize {\bf\textcolor{orange}{0.604}} & \footnotesize {\bf\textcolor{orange}{0.599}} &  \footnotesize {\bf\textcolor{orange}{0.57}}\\
\footnotesize C.2 & \footnotesize Wiki-727K & \footnotesize BOLT & \footnotesize Wiki-272K & \footnotesize 0.433 & \footnotesize 0.435 & \footnotesize 0.501 \\
\footnotesize C.3 & \footnotesize Wiki-727K & \footnotesize Topical Chat & \footnotesize Wiki-727K & \footnotesize 0.513 & \footnotesize 0.509 & \footnotesize 0.411 \\
\hline

\footnotesize C.4 & \footnotesize BOLT & \footnotesize - & \footnotesize Wiki-727K & \footnotesize 0.487 & \footnotesize  0.492 & \footnotesize 0.12 \\
\footnotesize C.5 & \footnotesize BOLT & \footnotesize Topical Chat & \footnotesize Wiki-727K & \footnotesize 0.489 & \footnotesize 0.478 & \footnotesize 0.027 \\
\hline

\footnotesize C.6 & \footnotesize Topical Chat & \footnotesize - &  \footnotesize Wiki-727K &  \footnotesize 0.505 &  \footnotesize 0.502 &  \footnotesize 0.020\\
\footnotesize C.7 & \footnotesize Topical Chat & \footnotesize BOLT & \footnotesize Wiki-727K & \footnotesize 0.5089 & \footnotesize 0.511 & \footnotesize 0.198\\
\hline
\end{tabular}}\\


\caption{ Effect of model architectures and training strategies on topic segmentation tasks, grouped by the test dataset of choice - Topical Chat {\bf(A.1 - 1.7)}, BOLT  {\bf(B.1 - B.7)}  \& Wiki-727K {\bf(C.1 - C.7)}. The best F1 scores for \textcolor{cyan}{\bf{BOLT}} (unstructured), \textcolor{purple}{\bf{Topical Chat}} (semi-structured), and \textcolor{orange}{\bf{Wiki-727K}}, per model are highlighted. Comparing the F1 scores in different pre-training \& fine-tuning scenarios, we conclude that the Cross-Segment BERT model {\it consistently} outperforms the Hierarchical Bi-LSTM, and CSRoBERTa model on {\it almost} all tasks. We also find that large-scale pre-training with the structured Wiki-727K dataset (and then fine-tuning with data from Target domain - underlined; Task B.2 vs. B.4 and Task A.3 vs. A.6) is not required to create cohesive topic segments on unstructured or semi-structured conversational data.(Train-finetuning-test split described in Section \ref{appdx:exp_setup}).}
\label{tab:exp4}

\end{table*}
\end{center}

\subsection{Effectiveness of Pre-training on Wiki-727K}
In this section, we investigate the effectiveness of pre-training on the large Wiki-727K dataset. We first pre-train the Hierarchical Bi-LSTM, CSRoBERTa, and CSBERT models on the 80\% of the Wiki-727K corpus. We further fine-tune the models on the unstructured datasets: BOLT chat and Topical Chat dataset. Additionally, we experiment with fine-tuning the models after training on semi-structured/unstructured dataset instead of the pre-training step with Wiki-727K.\\

From Table \ref{tab:exp4}, we find that training models with unstructured/semi-structured and fine-tuning it with semi-structured and unstructured dataset respectively, leads to better performance than fine-tuning with the Wiki-727K checkpoint. For instance, we find that with the pre-train with Wiki-727K and fine-tune paradigm, using the CSBERT, we obtain a F1 score of 0.725 for Topical Chat, whereas only training on Topical Chat results in a F1 score of 0.764 ({\it Task A.3 vs.A.6})\footnote{We find that for Cross-Segment BERT model Task A.6 and A.7 have very similar F1 scores}. For BOLT, the pre-train and fine-tune paradigm results in a F1 score of 0.489 whereas training from scratch with BOLT dataset results in a F1 score of 0.499 ({\it Task B.2 vs. B.4}). These results indicate that the conventional approach of pre-training on a large structured Wiki-727K dataset, and then fine-tuning with semi-structured or unstructured dataset, doesn't lead to an improvement in F1 scores, making the approach of pre-training on structured dataset questionable. We associate this finding to the fact that - Wiki-727K, although large enough for pre-training approaches and has been used in established Topic Segmentation methods for structured texts, the dataset fails to represent the rapid change in themes of conversations, thus making feature reuse \cite{neyshabur2020being} from the pre-training process redundant. In chats between two human agents, the topic of the conversation can change very quickly, which is not fully represented by feature hierarchy learned from structured texts.

\subsection{Effect of architecture on unstructured datasets}

We test our initial conclusions (Table \ref{tab:exp4}) on the efficacy of architectures across various number of segments. We find that, across different number of segments, the CSBERT model performs substantially better than the Hierarchical Bi-LSTM model and slightly better than CSRoBERTa model. From Figure  \ref{fig1}, we conclude that:

\begin{figure}{}{}
\hspace{-1.5cm}
    \centering
 	\includegraphics[width = 0.8\textwidth]{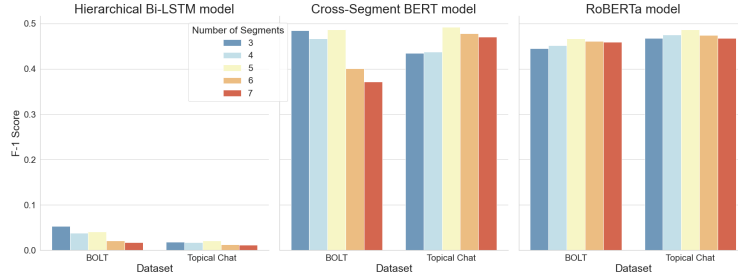}
 	\caption{ \footnotesize 
 	Effect of model architecture and the number of segments on segmentation performance. Comparing the F1 scores from all three models pre-trained on Wiki-727K and inferred on unstructured \& semi-structured datasets (without any fine-tuning), we find that CSBERT outperforms the other models robustly across the varying number of segments. Bi-LSTM performs very poorly compared to the other two models, which can be explained by its inability to fully consider the semantic context of the text representing segment boundaries. Additionally, with an increase in the number of segments, the F1 score peaks at 5 segments, and then drops across all models. We attribute this finding to the fact that at 5 segments, we are able to capture a relatively decent break in the theme of conversations for both datasets.} 
 	\label{fig1}
\end{figure}

\begin{itemize}[noitemsep,topsep=0pt,parsep=0pt,partopsep=0pt]
\item Across the varying number of segments curated and adapted during the pre-training step with Wiki-727K, the CSBERT model results in higher F1 scores when tested against semi-structured Topical Chat and unstructured BOLT datasets. Hence, we can conclude that for these topic segmentation tasks, CSBERT model is more suitable than the Hierarchical Bi-LSTM and the CSRoBERTa models.
\item As the number of segments in the conversational datasets increases, the F1 scores from all models drop. We associate this finding with the fact that the large segments of the conversational data may contain more heterogeneous topics, making it difficult for all models to group coherent chats.
\end{itemize}

\subsection{Practical recipes to improve unstructured segmentation tasks}

Casting topic segmentation for unstructured datasets as a binary classification problem leads to a severe imbalance in class labels, prompting the need to re-weight the samples or even modify the loss function to boost performance. The number of end-of-segment sentences (encoded as `1') is significantly smaller than the non-boundary sentences (sentences that do not mark the end of a segment; encoded as label `0') due to the inherent structure of segments in any document or chat. 

\subsubsection{Re-weighting in cross-entropy loss}:

To reduce the effect of dominance by the samples with label `0' and avoid biasing the model at inference time, we re-weight the class labels in cross-entropy loss function, giving proportional importance to labels `0' and `1'. The set of weights to be assigned is considered as a hyper-parameter and is optimized in the range {$[0, 1]$}.

We find that weighting the end-of-segment sentences (encoded as `1') with 0.8, and weighting the rest with 0.2 yields the best results on these datasets. From Table \ref{tab:exp10}, we conclude that re-weighting the cross-entropy loss function to provide proportional importance to both labels leads to a slightly better F1 score for all three models.

\subsubsection{Focal loss as an alternative loss function for imbalanced topic segmentation}\label{Focallossdescp}:

Focal loss has been used widely to mitigate the risks involved with class imbalance for tasks related to object detection \cite{lin2017focal}, credit-card fraud detection \cite{focalloss}, and other tasks involving class-imbalance \cite{mukhoti2020calibrating}. We consider the \(\alpha\) (a parameter that controls trade-off between precision and recall) and \(\gamma\) (focusing parameter; defines the degree of confidence assigned by the model to correct predictions that contributes to overall loss  values) as focal-loss hyper-parameters and tune these over 10 epochs.\\

Focal loss(\ref{second_eqn}) is different from Cross Entropy loss ( \ref{first_eqn}), as the former implements a technique called as "down-weighting", that reduces the influence of confidently predicting easy examples (predicted probability: $p >> 0.5$) on the loss function, resulting in more attention being paid to hard-to-predict examples (misclassified examples). To achieve this, an additional modulating-factor, called the focusing parameter ($\gamma$) is included to improve the conventional Cross Entropy loss function. Additionally, Focal loss also tackles the class-imbalance problem by introducing a weighting parameter ($\alpha$) to place appropriate weights on positive and negative classes.

\begin{equation}
    CE(p,y) =
\begin{cases}
    -log(p),& \text{if } y = 1\\
    -log(1-p),              & \text{otherwise}
\end{cases}
\label{first_eqn}
\end{equation}

\begin{equation}
    FL(p,y) = 
    \begin{cases}
    -\alpha(1-p)^{\gamma}log(p), & \text{if } y = 1 \\
    -(1-\alpha)p^{\gamma}log(1-p), & \text{otherwise}
    \end{cases}
\label{second_eqn}
\end{equation}

Thus, the Focal Loss function is a dynamically scaled Cross Entropy loss, where the scaling factor decays to zero as confidence in the correct class increases.\\

\begin{center}
\begin{table*}[h]
\scalebox{0.9}{
\hskip -1cm
\begin{tabular} {c|c|c|c|c|c|c|c|c}
\hline
\multicolumn{3}{c|}{\textbf{Datasets}}
&
\multicolumn{3}{c|}{\textbf{CE weights = [0.2, 0.8]}}

&
\multicolumn{3}{c}{ \textbf{Focal Loss ( \(\alpha\) = 0.8; \(\gamma\) = 2)}}
\\
\hline
\textbf{Pre-Train} 
&
\textbf{Finetune}
&
\textbf{Test}
&
\textbf{CSBERT}
&
\textbf{CSRoBERTa}
&
\textbf{Bi-LSTM}
&
\textbf{CSBERT}
&
 \textbf{CSRoBERTa}
&
\textbf{Bi-LSTM}
\\
\hline
Wiki-727K &   - &  Topical Chat &  0.497 & 0.491 &   0.028 &  0.512 &  0.504&   0.037\\
 Wiki-727K &  BOLT & Topical Chat & 0.475 &  0.411 &  0.398 &  0.483 &  0.427 &  0.405\\
 Wiki-727K &  Topical Chat &  Topical Chat & 0.729 &  0.717 & 0.933 &  0.748 &  0.729 &  0.928  \\
\hline
BOLT & - &  Topical Chat &  0.493 &  0.5 &  0.613 & 0.501&  0.510& 0.606\\
 BOLT &  Topical Chat &  Topical Chat &  0.736 &  0.731 &  0.917 & 0.747&   0.741&  0.920\\
\hline
 Topical Chat &  - &  Topical Chat &  0.767 &  {\bf\textcolor{purple}{0.769}} & {\bf\textcolor{purple}{0.952}} &  {\bf\textcolor{purple}{0.778}}& {\bf\textcolor{purple}{0.775}}& {\bf\textcolor{purple}{0.95}}\\
 Topical Chat & BOLT & Topical Chat & {\bf\textcolor{purple}{0.768}} & 0.761 & 0.505 & 0.777& 0.768& 0.515\\
\hline
\hline
 Wiki-727K &  - &  BOLT &  0.490 &  0.468 &  0.007 &  0.493 &  0.472 &  0.012\\
 Wiki-727K &  BOLT &  BOLT & 0.495 &  0.481 &  0.409 &   0.503 &  0.485 &  0.432\\
 Wiki-727K &  Topical Chat &  BOLT &  0.573 & 0.495 &  0.183 &  0.560 &  0.524 &  0.214\\
\hline

 BOLT & - &  BOLT &{\bf\textcolor{cyan}{0.575}} &{\bf\textcolor{cyan}{0.567}} & {\bf\textcolor{cyan}{0.443}} &   {\bf\textcolor{cyan}{0.580}} &  {\bf\textcolor{cyan}{0.572}}&  {\bf\textcolor{cyan}{0.45}}\\
 BOLT &  Topical Chat &  BOLT & 0.520 &  0.511 &  0.183 &   0.531 &   0.519&   0.189\\
\hline
 Topical Chat &  - &  BOLT &  0.546 &  0.544 &  0.159 &  0.555&  0.551&  0.164\\
 Topical Chat &  BOLT & BOLT &  0.537 & 0.531 &  0.333 &   0.549 & 0.54&  0.34\\

\hline

\hline

 Wiki-727K &  - &  Wiki-727K & {\bf\textcolor{orange}{0.609}} &  {\bf\textcolor{orange}{0.602}} &   {\bf\textcolor{orange}{0.591}} &  {\bf\textcolor{orange}{0.614}} &  {\bf\textcolor{orange}{0.611}} &   {\bf\textcolor{orange}{0.6}}\\
 Wiki-727K &  BOLT &  Wiki-272K &  0.435 &  0.438 &  0.508 &  0.441 & 0.446&  0.513 \\
 Wiki-727K &  Topical Chat &  Wiki-727K &  0.516 & 0.510 & 0.414 &  0.522 &  0.517 &   0.417\\
\hline
BOLT &  - &  Wiki-727K &  0.489 &   0.495 & 0.127 & 0.493&  0.501&   0.132\\
 BOLT &  Topical Chat &  Wiki-727K &  0.492 &  0.479 &  0.03 &   0.496&   0.484&  0.037\\
\hline

Topical Chat &  - &   Wiki-727K &  0.507 &  0.503 &  0.023 &  0.516&  0.510&  0.031\\
Topical Chat & BOLT & Wiki-727K &  0.511 &  0.514 & 0.199 &   0.520&  0.521&  0.210\\
\hline
\end{tabular}
}

\caption{Comparison of the F1 scores with different pre-training \& fine-tuning scenarios using re-weighting and Focal Loss. The best F1 scores for \textcolor{cyan}{\bf{BOLT}}, \textcolor{purple}{\bf{Topical Chat}}, and \textcolor{orange}{\bf{Wiki-727K}}, per model are highlighted. We see that Focal loss stands out to be a much a robust alternative than the re-weighted cross-entropy loss for topic segmentation of unstructured texts.}
\label{tab:exp10}

\end{table*}
\end{center}

We experiment with re-weighted cross-entropy loss and Focal loss functions for all combinations of pre-training and fine-tuning strategies and present our findings in Table \ref{tab:exp10}. We observed that, across {\it almost} (except 3) all paradigms, Cross-Segment BERT model outperforms Cross-Segment RoBERTa and Hierarchical Bi-LSTM model on both re-weighting and focal loss function recipes. Conversely, the F1 scores of Hierarchical Bi-LSTM are highly inconsistent. Hence, we can conclude the re-weighting of cross-entropy loss and replacement by focal loss function in our baseline model did not prove of large significance to the Bi-LSTM framework. We can attribute this finding to the fact that vanilla recurrent neural network architecture structure limits contextual learning around the segment boundaries, whereas Transformer-based architectures, with their multi-head attention mechanism, is able to capture the local context surrounding the boundaries of the segments more robustly, leading to a higher degree of topic coherence in the segments.\\

Moreover, Focal loss across all pre-training and fine-tuning strategies prove to be a robust alternative to re-weighting cross-entropy loss, for all three models. We hypothesize this effect to be a result of including appropriate values of focusing parameter($\gamma$) and trade-off parameter($\alpha$) in the Focal loss function, which assigns larger importance to hard-to-train examples, ensuring comprehensive learning of the latent feature hierarchy \cite{neyshabur2020being} of unstructured conversation data.


From Table \ref{tab:exp10}, we find that the resultant F1 scores with focal loss are higher than re-scaling the cross-entropy loss function. Hence, we conclude that for future iterations of topic segmentation tasks involving unstructured datasets, focal loss is a better alternative.

\section{Conclusion}

In this work, we evaluated the effectiveness of current and new Topic Segmentation methods on unstructured conversations. Our findings suggest that, across different model architectures  and datasets, pre-training on the large, structured Wiki-727K {\it is not required} for segmentation of unstructured conversational datasets such as Topical Chat and BOLT that contain syntactical and semantical noise. Surprisingly, training from scratch with only a few-examples provides a sufficiently strong baseline for this task. This finding challenges the prevalent pre-training on a large corpus and fine-tuning on the target domain paradigm, commonly used in a variety of tasks in current NLP research. Furthermore, we expanded upon existing Language Models by analyzing the effectiveness of Cross-Segment RoBERTa, which demonstrated better segmenting capabilities compared to the hierarchical Bi-LSTM model. Additionally, we provided various practical recipes to boost the topic segmentation performance on conversational and unstructured datasets, particularly achieved by using focal loss instead of cross-entropy loss. 

In conclusion, our research contributes valuable insights into the nuances of topic segmentation in conversational data and offers practical recommendations for achieving improved performance on these chat datasets. The findings presented here not only advance the understanding of the pre-training and fine-tuning paradigm, but also provide a solid foundation for further exploration and development of more effective topic segmentation techniques in the ever-evolving field of NLP.

\end{document}